\documentclass[11pt]{article}

\usepackage{acl07}
\usepackage{times}
\usepackage{latexsym}
\usepackage{haldefs}
\usepackage{psfig}
\usepackage{rotating}
\usepackage{multirow}
\usepackage{hvdashln}
\newcommand{\mysection}[1]{\vspace{-3mm}\section{#1}\vspace{-2mm}}
\newcommand{\mysubsection}[1]{\vspace{-2mm}\subsection{#1}\vspace{-1mm}}
\newcommand{\pitem}{\vspace{-2mm}\item}
\newcommand{\bibpinch}{\vspace{-1mm}}

\title{A Bayesian Model for Discovering Typological Implications}

\author{Hal Daum\'e III\\
        School of Computing\\
        University of Utah\\
        {\tt me@hal3.name} \And
        Lyle Campbell\\
        Department of Linguistics\\
        University of Utah\\
        {\tt lcampbel@hum.utah.edu}}

\date{}

\begin{document}
\maketitle

\begin{abstract}
A standard form of analysis for linguistic typology is the universal
implication.  These implications state facts about the range of extant
languages, such as ``if objects come after verbs, then adjectives come
after nouns.''  Such implications are typically discovered by
painstaking hand analysis over a small sample of languages.  We
propose a computational model for assisting at this process.  Our
model is able to discover both well-known implications as well as some
novel implications that deserve further study.  Moreover, through a
careful application of hierarchical analysis, we are able to cope with
the well-known sampling problem: languages are not independent.
\end{abstract}

\mysection{Introduction}

Linguistic typology aims to distinguish between logically possible
languages and actually observed languages.  A fundamental building
block for such an understanding is the \emph{universal implication}
\cite{greenberg63universals}.  These are short statements that
restrict the space of languages in a concrete way (for instance
``object-verb ordering implies adjective-noun ordering'');
\newcite{croft03typology}, \newcite{hawkins83wordorder} and
\newcite{song01typology} provide excellent introductions to linguistic
typology.  We present a statistical model for automatically
discovering such implications from a large typological database
\cite{wals}.

Analyses of universal implications are typically performed by
linguists, inspecting an array of $30$-$100$ languages and a few pairs
of features.  Looking at all pairs of features (typically several
hundred) is virtually impossible by hand.  Moreover, it is
insufficient to simply look at counts.  For instance, results
presented in the form ``verb precedes object implies prepositions in
16/19 languages'' are nonconclusive.  While compelling, this is not
enough evidence to decide if this is a statistically well-founded
implication.  For one, maybe $99\%$ of languages have prepositions:
then the fact that we've achieved a rate of $84\%$ actually seems
really bad.  Moreover, if the $16$ languages are highly related
historically or areally (geographically), and the other $3$ are not,
then we may have only learned something about geography.

In this work, we propose a statistical model that deals cleanly with
these difficulties.  By building a computational model, it is possible
to apply it to a very large typological database and search over many
thousands of pairs of features.  Our model hinges on two novel
components: a statistical noise model a hierarchical inference over
language families.  To our knowledge, there is no prior work directly
in this area.  The closest work is represented by the books
\emph{Possible and Probable Languages} \cite{newmeyer05probable} and
\emph{Language Classification by Numbers} \cite{mcmahon05language},
but the focus of these books is on automatically discovering
phylogenetic trees for languages based on Indo-European cognate sets
\cite{dyen92indoeuropean}.

\mysection{Data} \label{sec:data}

\begin{figure}[t]
\hspace{3mm}\psfig{figure=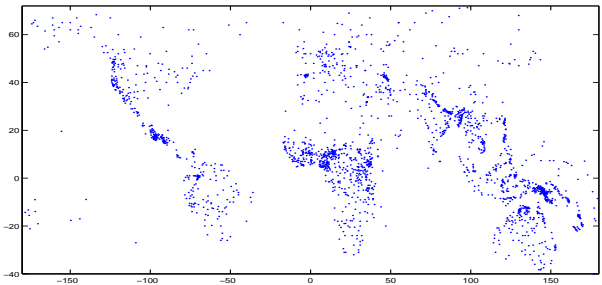,width=7.7cm,height=3.6cm}
\caption{Map of the $2150$ languages in the database.}\vspace{-4mm}
\label{fig:world}
\end{figure}

\begin{table*}[t]
\footnotesize
\begin{center}
\begin{tabular}{|l|c|c|c|c|c|c|}
\hline
& {\bf Numeral} & & & {\bf Glottalized} & & {\bf Number of} \\
{\bf Language} & {\bf Classifiers} & {\bf Rel/N Order} & {\bf O/V Order} & {\bf Consonants} & {\bf Tone} & {\bf Genders} \\
\hline
%7
%15
%27
%38
%44
%114
English     & Absent & NRel & VO & None & None & Three  \\
Hindi       & Absent & RelN & OV & None & None & Two  \\
Mandarin    & Obligatory & RelN & VO & None & Complex & None  \\
Russian     & Absent & NRel & VO & None & None & Three  \\
Tukang Besi & Absent & ? & Either & Implosives & None & Three  \\
Zulu        & Absent & NRel & VO & Ejectives & Simple & Five+  \\
\hline
\end{tabular}\vspace{-4mm}
\end{center}
\caption{Example database entries for a selection of diverse languages and features.}\vspace{-4mm}
\label{tab:example-features}
\end{table*}

The database on which we perform our analysis is the \emph{World Atlas
of Language Structures} \cite{wals}.  This database contains
information about $2150$ languages (sampled from across the world;
Figure~\ref{fig:world} depicts the locations of languages).  There are
$139$ \emph{features} in this database, broken down into categories
such as ``Nominal Categories,'' ``Simple Clauses,'' ``Phonology,''
``Word Order,'' etc.  The database is \emph{sparse}: for many
language/feature pairs, the feature value is unknown.  In fact, only
about $16\%$ of all possible language/feature pairs are known.  A
sample of five languages and six features from the database are shown
in Table~\ref{tab:example-features}.

Importantly, the density of samples is not random.  For certain
languages (eg., English, Chinese, Russian), nearly all features are
known, whereas other languages (eg., Asturian, Omagua, Frisian) that
have fewer than five feature values known.  Furthermore, some features
are known for many languages.  This is due to the fact that certain
features take less effort to identify than others.  Identifying, for
instance, if a language has a particular set of phonological features
(such as glottalized consonants) requires only listening to speakers.
Other features, such as determining the order of relative clauses and
nouns require understanding much more of the language.

\mysection{Models} \label{sec:models}

In this section, we propose two models for automatically uncovering
universal implications from noisy, sparse data.  First, note that even
well attested implications are not always exceptionless.  A common
example is that verbs preceding objects (``VO'') implies adjectives
following nouns (``NA'').  This implication (VO $\supset$ NA) has one
glaring exception: English.  This is one particular form of noise.
Another source of noise stems from transcription.  WALS contains data
about languages documented by field linguists as early as the 1900s.
Much of this older data was collected before there was significant
agreement in documentation style.  Different field linguists often had
different dimensions along which they segmented language features into
classes.  This leads to noise in the properties of individual
languages.

Another difficulty stems from the \emph{sampling problem.}  This is a
well-documented issue (see, eg., \cite{croft03typology}) stemming from
the fact that any set of languages is not sampled uniformly from the
space of all probable languages.  Politically interesting languages
(eg., Indo-European) and typologically unusual languages (eg.,
Dyirbal) are better documented than others.  Moreover,
languages are not independent: German and Dutch are more similar than
German and Hindi due to history and geography.

The first model, \textsc{Flat}, treats each language as independent.
It is thus susceptible to sampling problems.  For instance, the WALS
database contains a half dozen versions of German.  The \textsc{Flat}
model considers these versions of German just as statistically
independent as, say, German and Hindi.  To cope with this problem, we
then augment the \textsc{Flat} model into a \textsc{Hier}archical
model that takes advantage of known hierarchies in linguistic
phylogenetics.  The \textsc{Hier} model explicitly models the fact
that individual languages are \emph{not} independent and exhibit
strong familial dependencies.  In both models, we initially restrict
our attention to pairs of features.  We will describe our models as if
all features are binary.  We expand any multi-valued feature with $K$
values into $K$ binary features in a ``one versus rest'' manner.

\mysubsection{The \textsc{Flat} Model} \label{sec:models:flat}

In the \textsc{Flat} model, we consider a $2 \times N$ matrix of
feature values.  The $N$ corresponds to the number of languages, while
the $2$ corresponds to the two features currently under consideration
(eg., object/verb order and noun/adjective order).  The order of the
two features is important: $f_1$ implies $f_2$ is logically different
from $f_2$ implies $f_1$.  Some of the entries in the matrix will be
unknown.  We may safely remove all languages from consideration for
which \emph{both} are unknown, but we do \emph{not} remove languages
for which only one is unknown.  We do so because our model needs to
capture the fact that if $f_2$ is \emph{always} true, then $f_1
\supset f_2$ is uninteresting.

The statistical model is set up as follows.  There is a single
variable (we will denote this variable ``$m$'') corresponding to
whether the implication holds.  Thus, $m=1$ means that $f_1$ implies
$f_2$ and $m=0$ means that it does not.  Independent of $m$, we
specify two feature priors, $\pi_1$ and $\pi_2$ for $f_1$ and $f_2$
respectively.  $\pi_1$ specifies the prior probability that $f_1$ will
be true, and $\pi_2$ specifies the prior probability that $f_2$ will
be true.  One can then put the model together na\"ively as follows.
If $m=0$ (i.e., the implication does not hold), then the entire data
matrix is generated by choosing values for $f_1$ (resp., $f_2$)
independently according to the prior probability $\pi_1$ (resp.,
$\pi_2$).  On the other hand, if $m=1$ (i.e., the implication
\emph{does} hold), then the first column of the data matrix is
generated by choosing values for $f_1$ independently by $\pi_1$, but
the second column is generated differently.  In particular, if for a
particular language, we have that $f_1$ is true, then the fact that
the implication holds means that $f_2$ \emph{must} be true.  On the
other hand, if $f_1$ is false for a particular language, then we may
generate $f_2$ according to the prior probability $\pi_2$.  Thus,
having $m=1$ means that the model is significantly more constrained.
In equations:

\vspace{-3mm}
\begin{small}
\begin{align*}
p(f_1 \| \pi_1) &= \pi_1^{f_1} (1-\pi_1)^{1-f_1} \\
p(f_2 \| f_1, \pi_2, m) &=
  \brack{ f_2   & m = f_1 = 1 \\
          \pi_2^{f_2} (1-\pi_2)^{1-f_2} & \text{otherwise} }
%p(f_1=1 \| \pi_1) &= \pi_1 \\
%p(f_1=0 \| \pi_1) &= 1-\pi_1 \\
%p(f_2=1 \| f_1, \pi_2, m) &=
%   \brack{ 1     & m = f_1 = 1 \\
%           \pi_2 & \text{otherwise} }\\
%p(f_2=0 \| f_1, \pi_2, m) &=
%   \brack{ 0     & m = f_1 = 1 \\
%           1-\pi_2 & \text{otherwise} }
\end{align*}
\end{small}
\vspace{-3mm}

The problem with this na\"ive model is that it does not take into
account the fact that there is ``noise'' in the data.  (By noise, we
refer either to mis-annotations, or to ``strange'' languages like
English.)  To account for this, we introduce a simple noise model.
There are several options for parameterizing the noise, depending on
what independence assumptions we wish to make.  One could simply
specify a noise rate for the entire data set.  One could alternatively
specify a language-specific noise rate.  Or one could specify a
feature-specific noise rate.  We opt for a blend between the first and
second option.  We assume an underlying noise rate for the entire data
set, but that, conditioned on this underlying rate, there is a
language-specific noise level.  We believe this to be an appropriate
noise model because it models the fact that the majority of
information for a single language is from a single source.  Thus, if
there is an error in the database, it is more likely that other errors
will be for the same languages.

In order to model this statistically, we assume that there are latent
variables $e_{1,n}$ and $e_{2,n}$ for each language $n$.  If
$e_{1,n}=1$, then the first feature for language $n$ is wrong.
Similarly, if $e_{2,n}=1$, then the second feature for language $n$ is
wrong.  Given this model, the probabilities are exactly as in the
na\"ive model, with the exception that instead of using $f_1$ (resp.,
$f_2$), we use the exclusive-or\footnote{The exclusive-or of $a$ and
$b$, written $a \otimes b$, is true exactly when either $a$ or $b$ is
true but not both.} $f_1 \otimes e_1$ (resp., $f_2 \otimes e_2$) so
that the feature values are flipped whenever the noise model suggests
an error.

\begin{figure}[t]
\hspace{15mm}
\psfig{figure=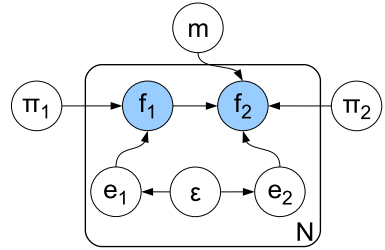,width=5cm}
\vspace{-3mm}
\caption{Graphical model for the \textsc{Flat} model.}
\vspace{-5mm}
\label{fig:flat-gm}
\end{figure}

The graphical model for the \textsc{Flat} model is shown in
Figure~\ref{fig:flat-gm}.  Circular nodes denote random variables and
arrows denote conditional dependencies.  The rectangular plate denotes
the fact that the elements contained within it are replicated $N$
times ($N$ is the number of languages).  In this model, there are four
``root'' nodes: the implication value $m$; the two feature prior
probabilities $\pi_1$ and $\pi_2$; and the language-specific error
rate $\ep$.  On all of these nodes we place Bayesian priors.  Since
$m$ is a binary random variable, we place a Bernoulli prior on it.
The $\pi$s are Bernoulli random variables, so they are given
independent Beta priors.  Finally, the noise rate $\ep$ is also given
a Beta prior.  For the two Beta parameters governing the error rate
(i.e., $a_\ep$ and $b_\ep$) we set these by hand so that the mean
expected error rate is $5\%$ and the probability of the error rate
being between $0\%$ and $10\%$ is $50\%$ (this number is based on an
expert opinion of the noise-rate in the data).  For the rest of the
parameters we use uniform priors.

%% The full hierarchical model is given below, where
%% ``$\dots$'' is the probabilistic model for $f_2$ specified above.

%% \begin{hierarchical}
%% m & & \Bin(\rho_m) \\
%% \pi_1 & & \Bet(a_1, b_1) \\
%% \pi_2 & & \Bet(a_2, b_2) \\
%% \ep   & & \Bet(a_\ep, b_\ep) \\
%% e_1   & \ep & \Bin(\ep) \\
%% e_2   & \ep & \Bin(\ep) \\
%% f_1   & e_1, \pi_1 & \Bin(\pi_1^{1-e_1} (1-\pi_1)^{e_1}) \\
%% f_2   & e_2, \pi_2, f_1, m & \dots
%% \end{hierarchical}

%% The values $\rho_m$, $a_1$, $b_1$, $a_2$, $b_2$, $a_\ep$ and $b_\ep$
%% are hyperparameters of the model.  

\mysubsection{The \textsc{Hier} Model} \label{sec:models:hier}

A significant difficulty in working with any large typological
database is that the languages will be sampled \emph{non}uniformly.
In our case, this means that implications that seem true in the
\textsc{Flat} model may only be true for, say, Indo-European, and the
remaining languages are considered noise.  While this may be
interesting in its own right, we are more interested in discovering
implications that are truly universal.

We model this using a hierarchical Bayesian model.  In essence, we
take the \textsc{Flat} model and build a notion of language
relatedness into it.  In particular, we enforce a hierarchy on the $m$
implication variables.  For simplicity, suppose that our ``hierarchy''
of languages is nearly flat.  Of the $N$ languages, half of them are
Indo-European and the other half are Austronesian.  We will use a
nearly identical model to the \textsc{Flat} model, but instead of
having a single $m$ variable, we have three: one for IE,
one for Austronesian and one for ``all languages.''

For a general tree, we assign one implication variable for each node
(including the root and leaves).  The goal of the inference is to
infer the value of the $m$ variable corresponding to the root of the
tree.

All that is left to specify the full \textsc{Hier} model is to specify
the probability distribution of the $m$ random variables.  We do this
as follows.  We place a zero mean Gaussian prior with (unknown)
variance $\si^2$ on the root $m$.  Then, for a non-root node, we use a
Gaussian with mean equal to the ``$m$'' value of the parent and tied
variance $\si^2$.  In our three-node example, this means that the root
is distributed $\Nor(0,\si^2)$ and each child is distributed
$\Nor(m_{\text{root}},\si^2)$, where $m_\text{root}$ is the random
variable corresponding to the root.  Finally, the leaves
(corresponding to the languages themselves) are distributed
\emph{logistic-binomial}.  Thus, the $m$ random variable corresponding
to a leaf (language) is distributed $\Bin(s(m_{\text{par}}))$, where
$m_\text{par}$ is the $m$ value for the parent (internal) node and $s$
is the sigmoid function $s(x) = [1 + exp(-x)]^{-1}$.

The intuition behind this model is that the $m$ value at each node in
the tree (where a node is either ``all languages'' or a specific
language family or an individual language) specifies the extent to
which the implication under consideration holds for that node.  A
large positive $m$ means that the implication is very likely to hold.
A large negative value means it is very likely to not hold.  The
normal distributions across edges in the tree indicate that we expect
the $m$ values not to change too much across the tree.  At the leaves
(i.e., individual languages), the logistic-binomial simply transforms
the real-valued $m$s into the range $[0,1]$ so as to make an
appropriate input to the binomial distribution.

\mysection{Statistical Inference}

In this section, we describe how we use Markov chain Monte Carlo
methods to perform inference in the statistical models described in
the previous section; \newcite{andrieu03mcmc} provide an excellent
introduction to MCMC techniques.  The key idea behind MCMC techniques
is to approximate intractable expectations by drawing random samples
from the probability distribution of interest.  The expectation can
then be approximated by an empirical expectation over these sample.

For the \textsc{Flat} model, we use a combination of Gibbs sampling
with rejection sampling as a subroutine.  Essentially, all sampling
steps are standard Gibbs steps, except for sampling the error rates
$e$.  The Gibbs step is not available analytically for these.  Hence,
we use rejection sampling (drawing from the Beta prior and accepting
according to the posterior).

The sampling procedure for the \textsc{Hier} model is only slightly
more complicated.  Instead of performing a simple Gibbs sample for $m$
in Step (4), we first sample the $m$ values for the internal nodes
using simple Gibbs updates.  For the leaf nodes, we use rejection
sampling.  For this rejection, we draw proposal values from the
Gaussian specified by the parent $m$, and compute acceptance
probabilities.

In all cases, we run the outer Gibbs sampler for $1000$ iterations and
each rejection sampler for $20$ iterations.  We compute the marginal
values for the $m$ implication variables by averaging the sampled
values after dropping $200$ ``burn-in'' iterations.

\mysection{Data Preprocessing and Search}

After extracting the raw data from the WALS electronic database
\cite{wals}\footnote{This is nontrivial---we are currently exploring
the possibility of freely sharing these data.}, we perform a minor
amount of preprocessing.  Essentially, we have manually removed
certain feature values from the database because they are
underrepresented.  For instance, the ``Glottalized Consonants''
feature has eight possible values (one for ``none'' and seven for
different varieties of glottalized consonants).  We reduce this to
simply two values ``has'' or ``has not.''  $313$ languages have no
glottalized consonants and $139$ have some variety of glottalized
consonant.  We have done something similar with approximately twenty
of the features.

For the \textsc{Hier} model, we obtain the hierarchy in one of two
ways.  The first hierarchy we use is the ``linguistic hierarchy''
specified as part of the WALS data.  This hierarchy divides languages
into families and subfamilies.  This leads to a tree with the leaves
at depth four.  The root has $38$ immediate children (corresponding to
the major families), and there are a total of $314$ internal nodes.
The second hierarchy we use is an areal hierarchy obtained by
clustering languages according to their latitude and longitude.  For
the clustering we first cluster all the languages into $6$
``macro-clusters.''  We then cluster each macro-cluster individually
into $25$ ``micro-clusters.''  These micro-clusters then have the
languages at their leaves.  This yields a tree with $31$ internal
nodes.

Given the database (which contains approximately $140$ features),
performing a raw search even over all possible \emph{pairs} of
features would lead to over $19,000$ computations.  In order to reduce
this space to a more manageable number, we filter:

\begin{small}
\begin{itemize}
\pitem There must be at least $250$ languages for which \emph{both}
  features are known.

\pitem There must be at least $15$ languages for which both feature
  values hold simultaneously.

\pitem Whenever $f_1$ is true, at least half of the languages also have
  $f_2$ true.
\end{itemize}
\end{small}

Performing all these filtration steps reduces the number of pairs
under consideration to $3442$.  While this remains a computationally
expensive procedure, we were able to perform all the implication
computations for these $3442$ possible pairs in about a week on a
single modern machine (in Matlab).

\mysection{Results}

The task of discovering universal implications is, at its heart, a
data-mining task.  As such, it is difficult to evaluate, since we
often do not know the correct answers!  If our model only found
well-documented implications, this would be interesting but useless
from the perspective of aiding linguists focus their energies on new,
plausible implications.  In this section, we present the results of
our method, together with both a quantitative and qualitative
evaluation.

\mysubsection{Quantitative Evaluation}

In this section, we perform a quantitative evaluation of the results
based on \emph{predictive power.}  That is, one generally would prefer
a system that finds implications that hold with high probability
across the data.  The word ``generally'' is important: this quality is
neither necessary nor sufficient for the model to be good.  For
instance, finding $1000$ implications of the form $A_1 \supset X, A_2
\supset X, \dots, A_{1000} \supset X$ is completely uninteresting if
$X$ is true in $99\%$ of the cases.  Similarly, suppose that a model
can find $1000$ implications of the form $X \supset A_1, \dots, X
\supset A_{1000}$, but $X$ is only true in five languages.  In both of
these cases, according to a ``predictive power'' measure, these would
be ideal systems.  But they are both somewhat uninteresting.

Despite these difficulties with a predictive power-based evaluation,
we feel that it is a good way to understand the relative merits of our
different models.  Thus, we compare the following systems:
\textsc{Flat} (our proposed flat model), \textsc{LingHier} (our model
using the phylogenetic hierarchy), \textsc{DistHier} (our model using
the areal hierarchy) and \textsc{Random} (a model that ranks
implications---that meet the three qualifications from the previous
section---randomly).

The models are scored as follows.  We take the entire WALS data set
and ``hide'' a random $10\%$ of the entries.  We then perform full
inference and ask the inferred model to predict the missing values.
The accuracy of the model is the accuracy of its predictions.  To
obtain a sense of the quality of the ranking, we perform this
computation on the top $k$ ranked implications provided by each model;
$k \in \{ 2,4,8,\dots,512,1024 \}$.

\begin{figure}[t]
\hspace{-2mm}
\psfig{figure=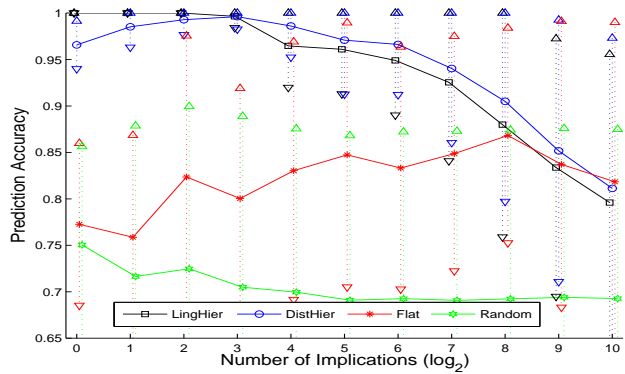,width=8.2cm,height=4.9cm}
\caption{Results of quantitative (predictive) evaluation.  Top curves
  are the hierarchical models; middle is the flat model;
  bottom is the random baseline.}
\label{fig:quant}
\end{figure}

The results of this quantitative evaluation are shown in
Figure~\ref{fig:quant} (on a log-scale for the x-axis).  The two
best-performing models are the two hierarchical models.  The flat
model does significantly worse and the random model does terribly.
The vertical lines are a standard deviation over $100$ folds of the
experiment (hiding a different $10\%$ each time).  The difference
between the two hierarchical models is typically not statistically
significant.  At the top of the ranking, the model based on
phylogenetic information performs marginally better; at the bottom of
the ranking, the order flips.  Comparing the hierarchical models to
the flat model, we see that adequately modeling the \emph{a priori}
similarity between languages is quite important.

\mysubsection{Cross-model Comparison}

The results in the previous section support the conclusion that the
two hierarchical models are doing something significantly different
(and better) than the flat model.  This clearly must be the case.  The
results, however, do not say whether the two hierarchies are
substantially different.  Moreover, are the results that they produce
substantially different.  The answer to these two questions is
``yes.''

We first address the issue of tree similarity.  We consider all pairs
of languages which are at distance $0$ in the areal tree (i.e., have
the same parent).  We then look at the mean tree-distance between
those languages in the phylogenetic tree.  We do this for all
distances in the areal tree (because of its construction, there are
only three: $0$, $2$ and $4$).  The mean distances in the phylogenetic
tree corresponding to these three distances in the areal tree are:
$2.9$, $3.5$ and $4.0$, respectively.  This means that languages that
are ``nearby'' in the areal tree are quite often very far apart in the
phylogenetic tree.

To answer the issue of whether the results obtained by the two trees
are similar, we employ Kendall's $\tau$ statistic.  Given two ordered
lists, the $\tau$ statistic computes how correlated they are.  $\tau$
is always between $0$ and $1$, with $1$ indicating identical ordering
and $0$ indicated completely reversed ordering.  The results are as
follows.  Comparing \textsc{Flat} to \textsc{LingHier} yield $\tau =
0.4144$, a very low correlation.  Between \textsc{Flat} and
\textsc{DistHier}, $\tau = 0.5213$, also very low.  These two are as
expected.  Finally, between \textsc{LingHier} and \textsc{DistHier},
we obtain $\tau=0.5369$, a very low correlation,
considering that both perform well predictively.

\mysubsection{Qualitative Analysis}

For the purpose of a qualitative analysis, we reproduce the top $30$
implications discovered by the \textsc{LingHier} model in
Table~\ref{tab:qualitative} (see the final page).\footnote{In truth,
our model discovers several tautalogical implications that we have
removed by hand before presentation.  These are examples like ``SVO
$\supset$ VO'' or ``No unusual consonants $\supset$ no glottalized
consonants.''  It is, of course, good that our model discovers these,
since they are obviously true.  However, to save space, we have
withheld them from presentation here.  The $30$th implication
presented here is actually the $83$rd in our full list.}  Each
implication is numbered, then the actual implication is presented.
For instance, \#7 says that any language that has adjectives preceding
their governing nouns also has numerals preceding their nouns.  We
additionally provide an ``analysis'' of many of these discovered
implications.  Many of them (eg., \#7) are well known in the
typological literature.  These are simply numbered according to
well-known references.  For instance our \#7 is implication \#18 from
Greenberg, reproduced by \newcite{song01typology}.  Those that
reference Hawkins (eg., \#11) are based on implications described by
\newcite{hawkins83wordorder}; those that reference Lehmann are
references to the principles decided by \newcite{lehmann81typology} in
Ch 4 \& 8.

Some of the implications our model discovers are obtained by
composition of well-known implications.  For instance, our \#3
(namely, OV $\supset$ Genitive-Noun) can be obtained by combining
Greenberg \#4 (OV $\supset$ Postpositions) and Greenberg \#2a
(Postpositions $\supset$ Genitive-Noun).  It is quite encouraging that
$14$ of our top $21$ discovered implications are well-known in the
literature (and this, not even considering the tautalogically true
implications)!  This strongly suggests that our model is doing
something reasonable and that there is true structure in the data.

In addition to many of the known implications found by our model,
there are many that are ``unknown.''  Space precludes attempting
explanations of them all, so we focus on a few.  Some are easy.
Consider \#8 (Strongly suffixing $\supset$ Tense-aspect suffixes):
this is quite plausible---if you have a language that tends to have
suffixes, it will probably have suffixes for tense/aspect.  Similarly,
\#10 states that languages with verb morphology for questions lack
question particles; again, this can be easily explained by an appeal
to economy.

Some of the discovered implications require a more involved
explanation.  One such example is \#20: labial-velars implies no
uvulars.\footnote{Labial-velars and uvulars are rare consonants (order
100 languages).  Labial-velars are joined sounds like /kp/ and /gb/ (to
English speakers, sounding like chicken noises); uvulars sounds are
made in the back of the throat, like snoring.}  It turns out that
labial-velars are most common in Africa just north of the equator,
which is also a place that has very few uvulars (there are a handful
of other examples, mostly in Papua New Guinea).  While this
implication has not been previously investigated, it makes some sense:
if a language has one form of rare consonant, it is unlikely to have
another.

As another example, consider \#28: Obligatory suffix pronouns implies
no possessive affixes.  This means is that in languages (like English)
for which pro-drop is impossible, possession is not marked
morphologically on the head noun (like English, ``book'' appears the
same regarless of if it is ``his book'' or ``the book'').  This also
makes sense: if you cannot drop pronouns, then one usually will mark
possession on the pronoun, not the head noun.  Thus, you do not need
marking on the head noun.

Finally, consider \#25: High and mid front vowels (i.e., /\:u/, etc.)
implies large vowel inventory ($\geq 7$ vowels).  This is supported by
typological evidence that high and mid front vowels are the ``last''
vowels to be added to a language's repertoire.  Thus, in order to get
them, you must also have many other types of vowels already, leading
to a large vowel inventory.

Not all examples admit a simple explanation and are worthy of further
thought.  Some of which (like the ones predicated on ``SV'') may just
be peculiarities of the annotation style: the subject verb order
changes frequently between transitive and intransitive usages in many
languages, and the annotation reflects just one.  Some others are
bizzarre: why not having fricatives should mean that you don't have
tones (\#27) is not a priori clear.

%% Some of the implications are harder to argue for (eg., \#20, \#23,
%% \#24, \#33) and would require further investigation.  We actually see
%% this as a positive result.  The entire purpose of this investigation
%% is to discover \emph{new} probable implications.  

\begin{table*}[t]
\centering
\small
\begin{tabular}{|rr@{ $\supset$ }l|l|}
\hline
{\bf \#} & {\bf Implicant} & {\bf Implicand} & {\bf Analysis} \\
\hline
 1 & Postpositions                   & Genitive-Noun              & Greenberg \#2a    \\
 2 & OV                              & Postpositions              & Greenberg \#4    \\
 3 & OV                              & Genitive-Noun              & Greenberg \#4 + Greenberg \#2a    \\
 4 & Genitive-Noun                   & Postpositions              & Greenberg \#2a (converse)    \\
 5 & Postpositions                   & OV                         & Greenberg \#2b (converse)    \\
\hline
 6 & SV                              & Genitive-Noun              & ???    \\
 7 & Adjective-Noun                  & Numeral-Noun               & Greenberg \#18    \\
 8 & Strongly suffixing              & Tense-aspect suffixes      & Clear explanation    \\
 9 & VO                              & Noun-Relative Clause       & Lehmann    \\
10 & Interrogative verb morph        & No question particle       & Appeal to economy    \\
\hline
11 & Numeral-Noun                    & Demonstrative-Noun         & Hawkins XVI (for postpositional languages)    \\
12 & Prepositions                    & VO                         & Greenberg \#3 (converse)    \\
13 & Adjective-Noun                  & Demonstrative-Noun         & Greenberg \#18    \\
14 & Noun-Adjective                  & Postpositions              & Lehmann    \\
15 & SV                              & Postpositions              & ???    \\
\hline
16 & VO                              & Prepositions               & Greenberg \#3 \\
17 & Initial subordinator word       & Prepositions               & Operator-operand principle (Lehmann)    \\
18 & Strong prefixing                & Prepositions               & Greenberg \#27b   \\
19 & Little affixation               & Noun-Adjective             & ???    \\
20 & Labial-velars                   & No uvular consonants       & See text    \\
\hline
21 & Negative word                   & No pronominal possessive affixes      & See text    \\
22 & Strong prefixing                & VO                         & Lehmann    \\
23 & Subordinating suffix            & Strongly suffixing         & ???    \\
24 & Final subordinator word         & Postpositions              & Operator-operand principle (Lehmann)    \\
25 & High and mid front vowels       & Large vowel inventories    & See text    \\
\hline
26 & Plural prefix                   & Noun-Genitive              & ???    \\
27 & No fricatives                   & No tones                   & ???    \\
28 & Obligatory subject pronouns     & No pronominal possessive affixes      & See text    \\
29 & Demonstrative-Noun              & Tense-aspect suffixes      & Operator-operand principle (Lehmann)     \\
30 & Prepositions                    & Noun-Relative clause       & Lehmann, Hawkins \\
\hline
\end{tabular}
\vspace{-2mm}
\caption{Top $30$ implications discovered by the \textsc{LingHier} model.}
\vspace{-4mm}
\label{tab:qualitative}
\end{table*}

\mysubsection{Multi-conditional Implications}

Many implications in the literature have multiple implicants.  For
instance, much research has gone into looking at which implications
hold, considering only ``VO'' languages, or considering only languages
with prepositions.  It is straightforward to modify our model so that
it searches over triples of features, conditioning on two and
predicting the third.  Space precludes an in-depth discussion of these
results, but we present the top examples in
Table~\ref{tab:multiconditional} (after removing the tautalogically
true examples, which are more numerous in this case, as well as
examples that are directly obtainable from
Table~\ref{tab:qualitative}).  It is encouraging that in the top
$1000$ multi-conditional implications found, the most frequently used
were ``OV'' ($176$ times) ``Postpositions'' ($157$ times) and
``Adjective-Noun'' ($89$ times).  This result agrees with intuition.

\begin{table}[t]
\centering
\small
\begin{tabular}{|r@{ }c@{ }l|}
\hline
{\bf Implicants} && {\bf Implicand} \\
\hline
Postpositions       & \multirow{2}{*}{$\supset$} & \multirow{2}{*}{Demonstrative-Noun} \\
Adjective-Noun      && \\
\hdashline
Posessive prefixes  & \multirow{2}{*}{$\supset$} & \multirow{2}{*}{Genitive-Noun} \\
Tense-aspect suffixes && \\              
\hdashline
Case suffixes       & \multirow{2}{*}{$\supset$} & \multirow{2}{*}{Genitive-Noun} \\
Plural suffix       && \\              
\hdashline
Adjective-Noun      & \multirow{2}{*}{$\supset$} & \multirow{2}{*}{OV} \\
Genitive-Noun       && \\              
\hdashline
High cons/vowel ratio & \multirow{2}{*}{$\supset$} & \multirow{2}{*}{No tones} \\
No front-rounded vowels && \\              
\hline
Negative affix      & \multirow{2}{*}{$\supset$} & \multirow{2}{*}{OV} \\
Genitive-Noun       && \\              
\hdashline
No front-rounded vowels        & \multirow{2}{*}{$\supset$} & \multirow{2}{*}{Large vowel quality inventory} \\
Labial velars       && \\              
\hdashline
Subordinating suffix  & \multirow{2}{*}{$\supset$} & \multirow{2}{*}{Postpositions} \\
Tense-aspect suffixes   && \\              
\hdashline
No case affixes     & \multirow{2}{*}{$\supset$} & \multirow{2}{*}{Initial subordinator word} \\
Prepositions        && \\              
\hdashline
Strongly suffixing  & \multirow{2}{*}{$\supset$} & \multirow{2}{*}{Genitive-Noun} \\
Plural suffix       && \\              
\hline
\end{tabular}
\caption{Top implications discovered by the \textsc{LingHier} multi-conditional model.}
\label{tab:multiconditional}
\end{table}

\mysection{Discussion}

We have presented a Bayesian model for discovering universal
linguistic implications from a typological database.  Our model is
able to account for noise in a linguistically plausible manner.  Our
hierarchical models deal with the sampling issue in a unique way, by
using prior knowledge about language families to ``group'' related
languages.  Quantitatively, the hierarchical information turns out to
be quite useful, regardless of whether it is phylogenetically- or
areally-based.  Qualitatively, our model can recover many well-known
implications as well as many more potential implications that can be
the object of future linguistic study.  We believe that our model is
sufficiently general that it could be applied to many different
typological databases --- we attempted not to ``overfit'' it to WALS.
Our hope is that the automatic discovery of such implications not only
aid typologically-inclined linguists, but also other groups.  For
instance, well-attested universal implications have the potential to
reduce the amount of data field linguists need to collect.  They have
also been used computationally to aid in the learning of unsupervised
part of speech taggers \cite{schone01univerals}.  Many extensions are
possible to this model; for instance attempting to uncover
typologically hierarchies and other higher-order structures.  We have
made the full output of all models available at
\url{http://hal3.name/WALS}.

\paragraph{Acknowledgments.}  We are grateful to Yee Whye Teh, Eric
Xing and three anonymous reviewers for their feedback on this work.

\begin{small}

\end{small}

%\begin{small}
%\bibliographystyle{acl}
%\bibliography{bibfile}

\begin{thebibliography}{}

\bibpinch
\bibitem[\protect\citename{Andrieu \bgroup et al.\egroup }2003]{andrieu03mcmc}
Christophe Andrieu, Nando de~Freitas, Arnaud Doucet, and Michael~I. Jordan.
\newblock 2003.
\newblock An introduction to {MCMC} for machine learning.
\newblock {\em Machine Learning (ML)}, 50:5--43.

\bibpinch
\bibitem[\protect\citename{Croft}2003]{croft03typology}
William Croft.
\newblock 2003.
\newblock {\em Typology and Univerals}.
\newblock Cambridge University Press.

\bibpinch
\bibitem[\protect\citename{Dyen \bgroup et al.\egroup
  }1992]{dyen92indoeuropean}
Isidore Dyen, Joseph Kurskal, and Paul Black.
\newblock 1992.
\newblock An {Indoeuropean} classification: A lexicostatistical experiment.
\newblock {\em Transactions of the American Philosophical Society}, 82(5).
\newblock American Philosophical Society.

\bibpinch
\bibitem[\protect\citename{Greenberg}1963]{greenberg63universals}
Joseph Greenberg, editor.
\newblock 1963.
\newblock {\em Universals of Languages}.
\newblock MIT Press.

\bibpinch
\bibitem[\protect\citename{Haspelmath \bgroup et al.\egroup }2005]{wals}
Martin Haspelmath, Matthew Dryer, David Gil, and Bernard Comrie, editors.
\newblock 2005.
\newblock {\em The World Atlas of Language Structures}.
\newblock Oxford University Press.

\bibpinch
\bibitem[\protect\citename{Hawkins}1983]{hawkins83wordorder}
John~A. Hawkins.
\newblock 1983.
\newblock {\em Word Order Universals: Quantitative analyses of linguistic
  structure}.
\newblock Academic Press.

\bibpinch
\bibitem[\protect\citename{Lehmann}1981]{lehmann81typology}
Winfred Lehmann, editor.
\newblock 1981.
\newblock {\em Syntactic Typology}, volume xiv.
\newblock University of Texas Press.

\bibpinch
\bibitem[\protect\citename{McMahon and McMahon}2005]{mcmahon05language}
April McMahon and Robert McMahon.
\newblock 2005.
\newblock {\em Language Classification by Numbers}.
\newblock Oxford University Press.

\bibpinch
\bibitem[\protect\citename{Newmeyer}2005]{newmeyer05probable}
Frederick~J. Newmeyer.
\newblock 2005.
\newblock {\em Possible and Probable Languages: A Generative Perspective on
  Linguistic Typology}.
\newblock Oxford University Press.

\bibpinch
\bibitem[\protect\citename{Schone and Jurafsky}2001]{schone01univerals}
Patrick Schone and Dan Jurafsky.
\newblock 2001
\newblock {\em Language Independent Induction of
Part of Speech Class Labels Using only Language Universals}.
\newblock Machine Learning: Beyond Supervision.

\bibpinch
\bibitem[\protect\citename{Song}2001]{song01typology}
Jae~Jung Song.
\newblock 2001.
\newblock {\em Linguistic Typology: Morphology and Syntax}.
\newblock Longman Linguistics Library.

\end{thebibliography}
%\end{small}

\end{document}